\documentclass[10pt,twocolumn,letterpaper]{article}

\usepackage{cvpr}
\usepackage{times}
\usepackage{epsfig}
\usepackage{graphicx}
\usepackage{amsmath}
\usepackage{amssymb}
\usepackage{pbox}
\usepackage{multirow}
\usepackage{bbm}

\usepackage{makecell}

\usepackage[T1]{fontenc}
\usepackage[utf8]{inputenc}
\usepackage{authblk}
\newcommand*{\affaddr}[1]{#1} 
\newcommand*{\affmark}[1][*]{\textsuperscript{#1}}
\newcommand*{\email}[1]{\texttt{#1}}

\usepackage[breaklinks=true,bookmarks=false]{hyperref}

\cvprfinalcopy 

\def\cvprPaperID{1226} 

 \ifcvprfinal\fi
\begin{document}

\title{A New Representation of Skeleton Sequences for 3D Action Recognition }

\author{%
Qiuhong~Ke\affmark[1], Mohammed~Bennamoun\affmark[1], Senjian~An\affmark[1], Ferdous~Sohel\affmark[2],    Farid~Boussaid\affmark[1]\\
\affaddr{\affmark[1]The University of Western Australia} ~~~
\affaddr{\affmark[2]Murdoch University}\\
\email{qiuhong.ke@research.uwa.edu.au}\\
\email{\{mohammed.bennamoun,senjian.an,farid.boussaid\}@uwa.edu.au}\\
\email{f.sohel@murdoch.edu.au}\\
}

\maketitle
   \thispagestyle{empty}

\begin{abstract}  

This paper presents a   new method for 3D action recognition with skeleton sequences
 (\ie,   3D trajectories of human skeleton joints).  The proposed method first transforms each skeleton sequence into three clips each consisting of several   frames  for spatial temporal feature    learning using  deep  neural  networks. 
Each clip is generated  from one channel of the   cylindrical coordinates of the skeleton sequence. Each frame of the generated clips represents the temporal information of  the entire skeleton sequence,
and incorporates one particular
spatial relationship between the joints. The entire clips include
multiple frames with different spatial relationships,
which provide useful spatial structural information of the human skeleton.
 We propose to use deep convolutional neural networks to learn long-term temporal information
of the skeleton sequence from the  frames of the generated clips, and then use a  Multi-Task Learning Network (MTLN)  to jointly process    all frames of the generated clips in parallel to incorporate spatial structural information   for action recognition. 
         Experimental results  clearly show the effectiveness of the proposed     new representation and  feature learning method for 3D action recognition.
\end{abstract}

\section{Introduction}
\label{intro}

  3D skeleton data records the trajectories of human skeleton joints and is robust to illumination changes and invariant to camera views \cite{han2016space}.
With the prevalence of highly-accurate and affordable devices, action recognition based on
3D skeleton sequence has been attracting increasing attention \cite{xia2012view, vemulapalli2014human, du2015hierarchical, Shahroudy_2016_CVPR, zhu2016co, liu2016spatio, koniusz2016tensor, wang2016action, ke2017skeletonnet}. In this paper, we focus on   
skeleton-based 3D action recognition. 

To recognize a video action, the  temporal information of the   sequence
 needs to be exploited  to understand the dynamics of human postures \cite{niebles2010modeling, gaidon2013temporal, wang2014latent,
fernando2015modeling, ke2016human}. For skeleton data, the spatial structure of the human skeleton is also an   important clue for action recognition \cite{zhu2016co}. Each skeleton sequence provides only the trajectory of human skeleton 
joints. The time series of the joints can be used  in  
  recurrent neural networks (RNNs) with Long-Short
Term Memory (LSTM) neurons \cite{graves2012neural, graves2013speech}    to explore the spatial structure and temporal structure of the skeleton sequence for action recognition \cite{du2015hierarchical, veeriah2015differential, zhu2016co, Shahroudy_2016_CVPR, liu2016spatio}.    Although LSTM networks are designed to explore the long-term temporal dependency problem, it is still difficult for LSTM to memorize the information
 of the entire sequence with many timesteps \cite{weston2014memory, gu2016recurrent}. In addition, it is also difficult to construct deep LSTM to extract 
 high-level features \cite{sainath2015convolutional, pascanu2013construct}.  
 
 Convolutional neural networks (CNNs) \cite{lecun1995convolutional}  
 nowadays  have achieved great success in 
  image classification \cite{chatfield2014return, ciregan2012multi, krizhevsky2012imagenet, simonyan2014very, szegedy2015going, xiong2015recognize, ke2014rotation}.  However, for video action recognition, it lacks the capacity to
  model the long-term temporal dependency of the entire video \cite{wang2016temporal}. 
   In this paper, instead of directly   exploring the long-term temporal information from the   skeleton sequences,  we first represent  the skeleton sequences as clips consisting of only a few frames. With the generated clips, the long-term temporal structure of the skeleton sequence can be effectively learned by using deep CNNs to process the frame images of the generated clips. In addition, the spatial structural information of the human skeleton can be exploited from the entire clips.
    
 More specifically, for each  skeleton sequence, we generate three clips corresponding to the three channels of the
  cylindrical coordinates of the skeleton sequence. Each clip consists of four frames, which are   generated   by computing the relative positions
  of   the joints to four reference joints.  
  Each frame of the clips   describes the  temporal information of  the entire skeleton sequence, and  includes    one particular spatial relationship between the joints. The entire clips  aggregate multiple frames with    different spatial relationships, providing important information of the spatial structure of the skeleton joints.

\begin{figure*}
\vspace{-1mm} 
\begin{center}
   \includegraphics[width=6.5in]{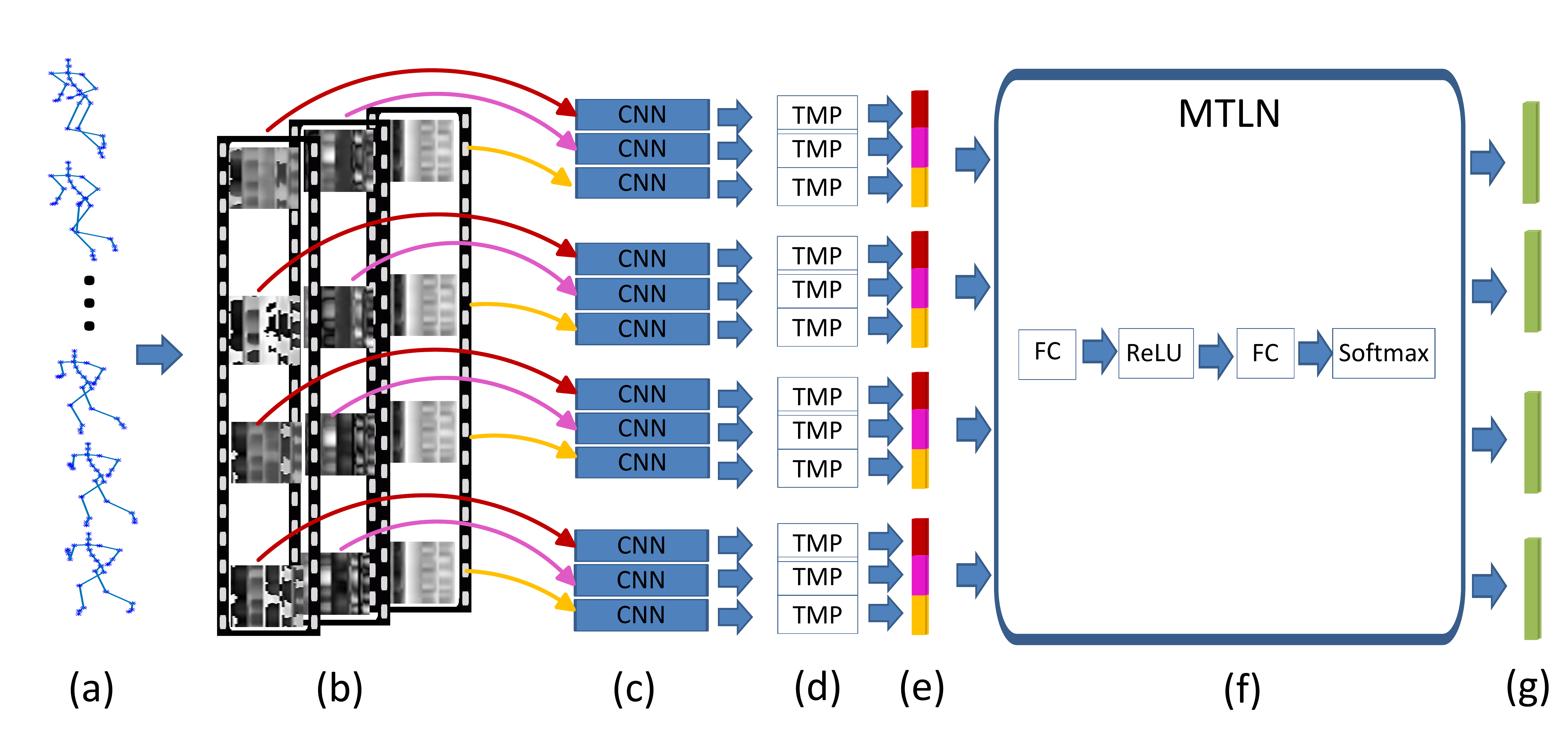}
\end{center}
\vspace{-1mm}
  \caption{Architecture of the proposed method. Given  a   skeleton sequence (a), three    clips (b)   corresponding to the three channels of the    cylindrical coordinates  are generated.  
   A deep  CNN model (c) and a  temporal mean pooling (TMP) layer (d)  are used to extract a compact representation from each frame of  the clips (see Figure \ref{cnn2} for details).    
    The output   CNN representations  of the three clips  at the same time-step are concatenated, resulting  four feature vectors (e). 
   Each feature vector represents  the temporal information of the skeleton sequence  and a  particular spatial relationship of the skeleton joints. 
   The proposed MTLN (f) which includes a fully connected (FC) layer,   a rectified linear unit  (ReLU), another FC layer and a Softmax layer  jointly processes the four feature vectors in parallel and     outputs four sets of class scores (g), each corresponding to one task of classification using one feature vector.
   During training, the loss values of the  four tasks are summed up  to define the loss value of the network   used to update the   network parameters. For testing, the class scores of the  four   tasks    are averaged to generate the final prediction of the action class.  } 
  
\label{cnn1}  
 
 \vspace{-1mm}
  \end{figure*}

  
 Since the temporal information of a skeleton sequence is  incorporated in the  frames   of the generated clips, 
 the long-term temporal structure of the skeleton sequence can be learned by extracting features from the frames 
 of the generated clips. More specifically, each frame of the generated clips is fed to a deep CNN to extract a CNN feature.
Then the three  CNN features of   the three clips at  the same time-step (See Figure \ref{cnn1}) are concatenated into one feature vector. 
  Consequently,  four feature vectors    are extracted from all the time-steps. 
  Each feature vector represents the temporal information of the skeleton sequence and one particular spatial relationship between the joints. 
 The  feature vectors of different time-steps represent    different spatial relationships  with intrinsic 
relationships among them.
This paper proposes to utilize the intrinsic relationships among different feature vectors for action recognition using a Multi-Task Learning Network (MTLN).
 Multi-task learning 
 aims at improving 
the generalization performance  by jointly training multiple related tasks and utilizing their intrinsic relationships \cite{caruana1998multitask}.
In the proposed MTLN, the classification of each feature vector  is treated as a separate task, and the MTLN  jointly learns  multiple classifiers each from one feature vector and outputs  multiple    predictions,  each corresponding to one task.
All the  feature vectors of the same skeleton sequence have the same label as the skeleton sequence. 
During training,  the loss value of each task is individually computed using its own   class scores. Then the  loss values of all tasks are      summed up to define the total loss of the network which is then used to  learn the network parameters. 
  During testing, the   class scores of all tasks are averaged to form the final prediction of the action class.   Multi-task learning simultaneously    solves multiple tasks with weight sharing, which can improve the performance 
 of individual tasks \cite{caruana1998multitask}.

The main contributions of this paper are summarized as follows. \textbf{(1)} We propose to transform each skeleton sequence to a new representation, \ie, three   clips, to allow global long-term temporal modelling of the skeleton sequence by using deep CNNs to learn hierarchical features from frame images.
  \textbf{(2)} We introduce a MTLN to process all the CNN features of the frames  in the generated clips, thus to learn the spatial structure and the temporal information of the skeleton sequence.   The MTLN improves the performance by   utilizing intrinsic relationships  among different frames of the generated clips. 
Our experimental results demonstrate that MTLN performs   better than   concatenating or pooling the features of the frames (See Section \ref{exp}).   \textbf{(3)} The proposed method   achieves the state-of-the-art performance on three skeleton datasets, including the large  scale NTU RGB+D dataset \cite{Shahroudy_2016_CVPR}.

\section{Related Works}
In this section, we cover the relevant    literature of    skeleton-based action recognition methods using  hand-crafted features
or using deep learning networks.
 
\textbf{Hand-crafted Features}~~
In \cite{hussein2013human},    
the covariance matrices  of the trajectories of the joint positions  are   computed over hierarchical temporal levels to model the skeleton sequences. In \cite{wang2012mining},  the pairwise relative positions of each joint with other joints are computed to represent each frame of the skeleton sequences, and Fourier Temporal Pyramid (FTP) is used to model the    temporal patterns. In \cite{yang2012eigenjoints},  the pairwise relative positions of the joints are also used to characterize posture features, motion features, and
offset features of the skeleton sequences.  Principal Component Analysis (PCA) is then applied to the normalized features to compute EigenJoints as  representations. In \cite{xia2012view},  histograms of 3D joint locations are computed to represent each frame  of the skeleton sequences, and  HMMs are   used to model the temporal dynamics.  In \cite{vemulapalli2014human},    the  rotations and translations   between various body parts are used as representations, and 
a skeleton sequence is modelled as a curve in the Lie group. The   temporal dynamics are modelled with     FTP.   

\textbf{Deep Learning Methods }~~
In \cite{du2015hierarchical}, the   skeleton joints are divided 
into five sets  corresponding to five body parts.  They are fed into five LSTMs  for feature fusion and classification.
In \cite{zhu2016co}, the skeleton joints are fed  to a deep LSTM 
at each time slot   to learn 
 the inherent co-occurrence features of skeleton joints.
  In \cite{Shahroudy_2016_CVPR}, the long-term context representations   of the body parts are learned with a part-aware LSTM.
  In \cite{liu2016spatio}, both the spatial and temporal information of skeleton sequences are  learned 
  with  a spatial temporal LSTM. A Trust Gate is also proposed to remove noisy joints. This method achieves the state-of-the-art performance on the   NTU RGB+D dataset \cite{Shahroudy_2016_CVPR}.

\section{Proposed Method}

An overall architecture of the proposed method is shown in Figure \ref{cnn1}. 
The proposed method starts by generating clips of skeleton sequences. 
 A skeleton sequence of any length  is 
 transformed into three   clips each consisting of several gray images. The generated clips are then fed to a deep CNN model to extract CNN features which are used  in a MTLN  for  action recognition.  
   
\subsection{Clip Generation}
 
\label{imagegene}

   \begin{figure} [h!]
\vspace{-3mm} 
\begin{center}
   \includegraphics[width=3.2in]{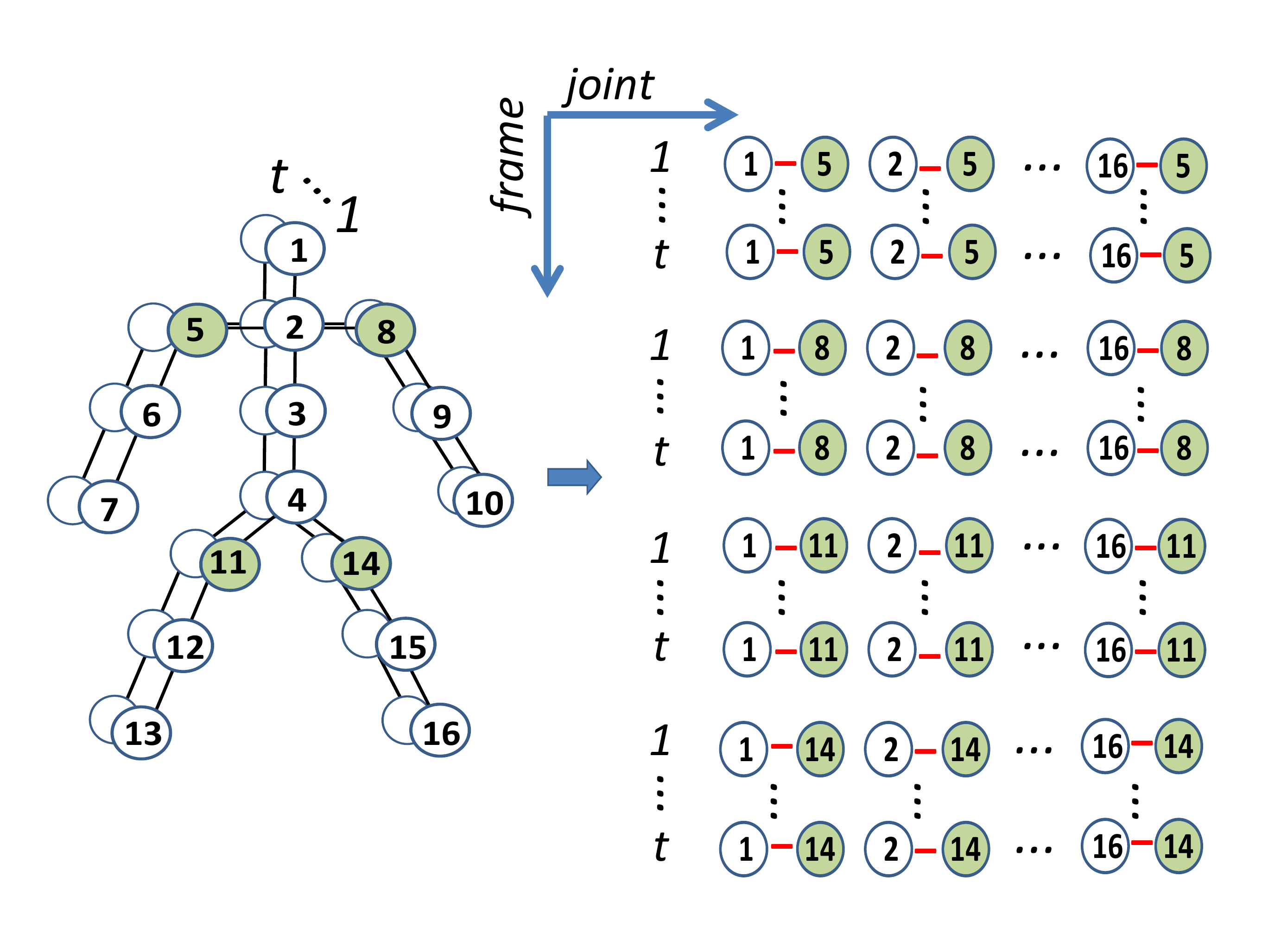}
\end{center}
\vspace{-3mm}
  \caption{Clip Generation of a skeleton sequence. The skeleton joints of each frame  are first  arranged as a chain by concatenating the joints of each body part (\ie, 1-2-3-...-16).  Four reference joints shown in green (\ie,  left shoulder 5, right shoulder 8, left hip 11 and right hip 14) are then respectively used to compute relative positions of  the other joints   to incorporate different spatial relationships between the joints. Consequently, four 2D arrays  are obtained   by combining the   relative positions of all the frames of the skeleton sequence. The relative position of each joint in the 2D arrays is   described with cylindrical coordinates. 	The four 2D arrays  corresponding to the same channel of the coordinates are transformed to four gray images and as a clip. Thus three clips are generated from the three channels of the cylindrical coordinates of the four 2D arrays.} 
  
\label{clip1}  
 
 \vspace{-1mm}
  \end{figure}

Compared to RGB videos which consist of multiple frame images,    skeleton sequences only  provide    the   trajectories of the 3D coordinates.   This paper proposes to transform the original skeleton sequence to    a collection of  clips each consisting of several images, thus to allow    spatial temporal feature learning using deep neural networks.  Intuitively, one could represent the content of  each frame of the skeleton sequence as an image to generate a video.
However, if the skeleton sequence has many frames, this method will result in a long video of which   the temporal dynamics will be difficult to learn. In addition,  
each frame of the generated video will also be very sparse as  the  number of the skeleton joints is small.  
To overcome this problem, we 	propose to represent the temporal dynamics of the skeleton sequence in a frame image, and then use multiple frames to incorporate    different spatial  relationships between the joints. An advantage of this method is that for any skeleton sequence of any length, the generated clips  contain  the same number of frames and the long-term temporal information of the original skeleton sequence can be effectively captured with the powerful  CNN   representations of the frame images in the generated clips. 
 
  As shown in Figure \ref{clip1}, for a skeleton sequence,  the skeleton joints of each frame are first arranged as a  chain by concatenating the joints of each body part. 
  Considering that the relative positions between joints  provide more useful information than their absolute locations (\eg, the relative location of the hand to the shoulder in ``pushing''), four reference joints, 
 namely, the left shoulder, the right shoulder, the left hip and the right hip,  are   respectively 
 used to compute relative positions  of  the other  joints, thus to incorporate different spatial relationships between joints and provide useful   structural information of the skeleton.   
  These four joints are selected as reference joints due to the fact that they are    stable in most actions. They can thus   reflect the motions of the other joints.  Although the base of the spine is also stable, it is close to the left hip and the right hip.  It is therefore discarded to avoid   information redundancy. By combing the relative joints   of all the frames, four 2D arrays  with dimension $(m-1)\times t$ are generated ($m $ is the number of skeleton joints in each frame and    $t$ is the number of  frames of the skeleton sequence). 
   The relative positions of joints in the   2D arrays are originally described with  3D Cartesian coordinates.  Considering that the cylindrical coordinates are more useful   to analyse the motions as each human body utilizes pivotal joint movements to perform an action, the 3D Cartesian coordinates are  transformed to cylindrical coordinates  in the proposed representation of skeleton sequences. The cylindrical coordinates have been used to extract view-invariant motion features for action recognition in \cite{weinland2006free}.
    The four 2D arrays  corresponding to the same channel of the 3D cylindrical coordinates are transformed to four gray images  by scaling the coordinate values between 0 to 255 using a linear transformation. A clip is then constructed with the four gray images.  Consequently, three clips are generated from the three channels of the 3D coordinates of the four 2D arrays. 
    

   \subsection{Clip Learning}

Each frame of the generated clips  describes the temporal dynamics of all frames of the skeleton sequence  and  one   particular spatial relationship between the skeleton joints in one channel of the cylindrical coordinates. 
 Different frames of the generated clip describe different spatial relationships and   
  there exists   intrinsic 
relationships among them.  
A deep CNN is first leveraged to extract a  compact representation from each frame of the generated clips to exploit the long-term temporal information of the skeleton sequence.
  Then the CNN features of  all frames of the generated clips are jointly processed in parallel using  multi-task learning,    thus to  utilize their intrinsic relationships to learn the spatial temporal information for 3D action recognition.

 \subsubsection{Temporal Pooling of CNN Feature Maps}
 \label{cnnrep}

To learn the features of the generated clips, a deep CNN is firstly employed  to extract a compact representation  of each frame of the clips.
Since each frame  describes the temporal dynamics of the   skeleton sequence,
 the spatial invariant CNN feature of each frame could thus represent the robust temporal information of the skeleton sequence.

Given the generated clips, the CNN feature of each frame is extracted with the pre-trained       VGG19 \cite{simonyan2014very}   model.
The pre-trained CNN model is leveraged as a feature extractor due to the fact that  the  CNN features extracted by the models pre-trained  with   ImageNet \cite{russakovsky2015imagenet} are very powerful and have been successfully applied in a number of cross-domain applications \cite{donahue2014decaf,girshick2014rich,sharif2014cnn,han2015matchnet}. In addition, 
 current skeleton datasets are either too small or too noisy to suitably train a deep network. 
Although the frames of the generated clips are not natural images, they could still be fed to the CNN model pre-trained with ImageNet \cite{russakovsky2015imagenet} for feature extraction.  
The similarity between a natural image and the generated frames is that both of them are matrices with some patterns.  The CNN models trained on the large image dataset can be used as a feature extractor  to extract  representations of the patterns in      matrices. The learned representations are generic and   can be transferred to novel tasks from the original tasks \cite{yosinski2014transferable, long2015learning}.

   \begin{figure}
\vspace{-1mm} 
\begin{center}
   \includegraphics[width=3.2in]{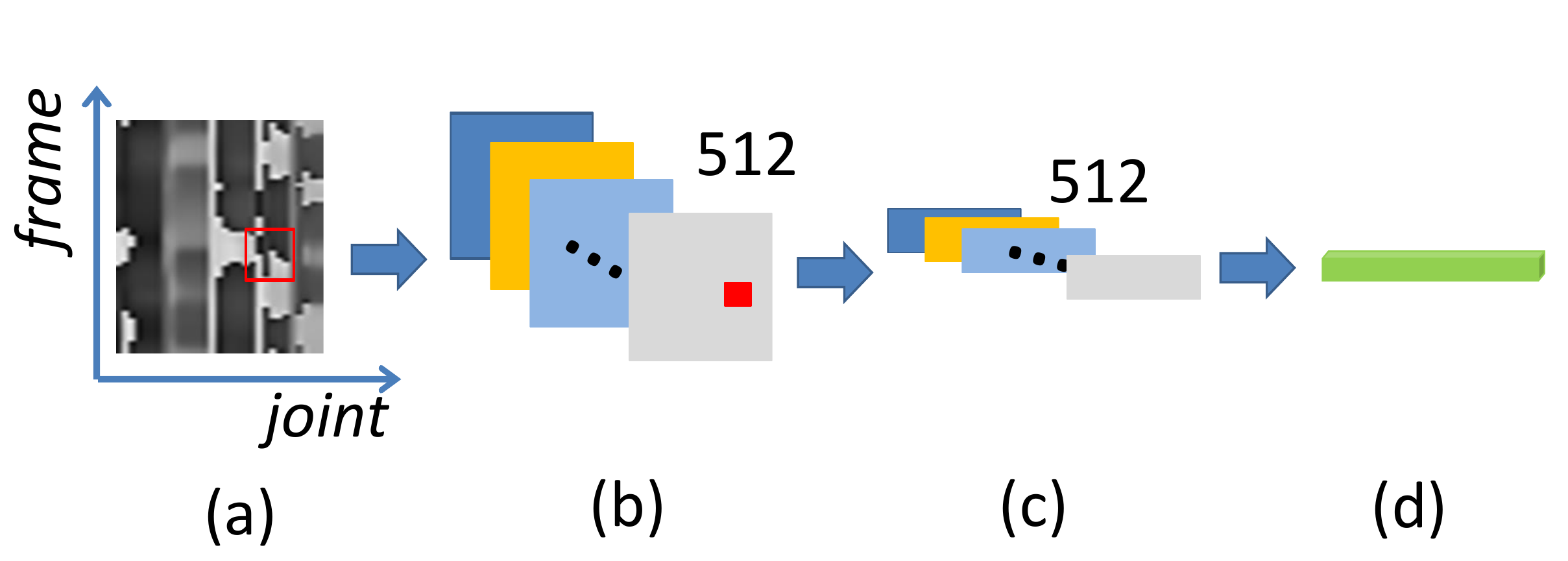}
\end{center}
\vspace{-1mm}
  \caption{\small{Temporal mean pooling   of the CNN feature maps. (a) An input frame of the generated clips, for which the rows   correspond to the different frames of the skeleton sequence and the  columns correspond to the different vectors generated from the joints. (b)   Output feature maps of the conv5\_1 layer. The size is  $14\times 14 \times 512$.  Each activation (shown in red)  of the feature map is  a    feature correspond to the local region of the original image (shown with a red square). 
(c) Temporal features of all joints of the skeleton sequence, which are obtained by applying  mean pooling  to    each feature map   in the row (temporal) dimension. 
(d) Output feature, which is achieved by concatenating all the feature maps in (c). 
}}
\label{cnn2}  
 
 \vspace{-1mm}
  \end{figure}

The pre-trained       VGG19 \cite{simonyan2014very}   model contains 5 sets of convolutional layers conv1, conv2,
..., conv5. Each set includes a stack of 2 or 4  convolutional layers with the same kernel size.
Totally there are  16 convolutional layers and 
three fully connected layers in the network.  
Although deep neural networks are capable of learning powerful and 
generic features which can be used in other novel domains,    the features extracted from the different layers have different transferability. Particularly, the features in earlier layers are more generic, while in later layers, the features are more task-specific,  which largely rely on the original classes and dataset. The features of the later layers are thus less suitable than those of the earlier layers to transfer to other domains \cite{yosinski2014transferable, long2015learning}.  
   Therefore, 
     this paper adopts a compact representation that is derived from the activations of the convolutional layer to exploit the   temporal information of  a skeleton sequence.  The feature maps in the convolutional layer have been successfully applied for action recognition and image retrieval \cite{peng2015encoding, radenovic2016cnn}.           
Specifically, the last 3 convolutional layers and fully connected layers of the network are discarded. 
Each frame image of the three clips is scaled to $224 \times 224$, and  is then duplicated three times to formulate a color image, so that it can be fed to the network. 
 The output of the   convolutional layer   conv5\_1 is used as the representation of the input frame, 
which   is a 3D tensor with size $14\times 14\times 512$, 
\ie,  512 feature maps with size $14\times 14$.  

The  rows  of the generated frame correspond to  different frames  of a skeleton sequence. The dynamics of the row features of the generated image therefore represents the temporal evolution of the skeleton sequence.
Meanwhile,    the activations of each feature map in the conv5\_1 layer are the local features corresponding to the local regions in the original input image \cite{peng2015encoding}. 
 The temporal information  of the sequence can thus be extracted from the   row features of the feature maps.
  More specifically, the feature maps   are processed with   temporal mean pooling  with kernel size $14\times 1$, \ie, the pooling is applied over the temporal, or row dimension, thus to generate a compact fusion representation from all temporal stages of the skeleton sequence. 
Let the activation at the $i^{th}$ row and the $j^{th}$ column of the  $k^{th}$ feature map   be $x_{i,j}^k$. After temporal mean pooling, the output of the $k^{th}$ feature map  is given by:
\begin{equation}
\begin{array}{c}
 \mathbf{y^k}=\left[y_1^k,\cdots, y_j^k, \cdots,y_{14}^k\right] \\
 \\
 y_j^k= \dfrac{1}{14}\sum\limits_{i=1}^{14}\max(0,x_{i,j}^k)\\
\end{array}
\end{equation}

The outputs of all feature maps (512) are concatenated to form a 7168D  ($14\times512=7168$) feature vector,  which  represents  the temporal dynamics of the skeleton sequence in one channel of the cylindrical coordinates.

 \subsubsection{Multi-Task Learning Network (MTLN)}

As shown in Figure \ref{cnn1}(e), the three   7168D features   of the three   clips at the same time-step are concatenated  to form a feature vector, generating four feature vectors in total. Each feature vector represents the temporal dynamics of the skeleton sequence and includes one particular spatial relationship between the joints in one of three  cylindrical coordinates. 
The four feature vectors have   intrinsic relationships between each other. 
 An MTLN is then proposed to jointly process the four   feature vectors to utilize their intrinsic relationships for action recognition. 
 The classification of each feature vector  is treated as a separate task with the same classification label  of the   skeleton sequence.
   
   The architecture of the network is shown in Figure  \ref{cnn1}(f). 
 It includes two fully connected (FC) layers and a  Softmax layer. 
Between the two FC layers there is a rectified linear unit (ReLU) \cite{nair2010rectified} to introduce an additional non-linearity.
   Given the four features as inputs,  the MTLN generates four  frame-level predictions,  each corresponding to one task.  During training,   the class scores of each task   are   used to compute a loss value. Then the  loss values of all tasks are      summed up to generate the final loss of the network used to  learn the network parameters. 
  During testing, the   class scores of all tasks are averaged to form the final prediction of the action class. The loss value of the  $k^{th}$ task    ($k=1,\cdots,4$) is given by Equation \ref{eql1}.

    \begin{equation}
  \label{eql1}
  \begin{array}{ll}
  \ell_k(\mathbf{z_k},\mathbf{y})&=\sum\limits_{i=1}^{m} y_i \left(-log\left(\dfrac{\exp z_{ki}}{\sum\limits_{j=1}^m \exp z_{kj}}\right)\right) \\
&=\sum\limits_{i=1}^{m} y_i \left(log\left(\sum\limits_{j=1}^m \exp z_{kj} \right)-z_{ki}\right)
  \end{array}
\end{equation}
where $\mathbf{z_k}$ is the   vector   fed to the Softmax layer generated from the $k^{th}$ input feature, $m$ is the number of action classes and $y_i$ is the ground-truth label for class $i$.
The final loss value of the network is   computed as the   sum of the
four individual losses, as shown below in Equation \ref{eql2}:

\begin{equation}
 \label{eql2}
\mathcal{L}(Z,\mathbf{y})=\sum\limits_{k=1}^{4} \ell_k (\mathbf{z_k},\mathbf{y})
\end{equation}
 where $Z=[\mathbf{z_1},\cdots,\mathbf{z_4}]$.

\section{Experiments and Analysis}

 The proposed method is tested on three skeleton action datasets: NTU RGB+D dataset \cite{Shahroudy_2016_CVPR}, SBU kinect interaction dataset \cite{yun2012two} and CMU dataset \cite{cmu2013}.

 The main ideas  of the proposed method  \textbf{Clips + CNN + MTLN } are 1) generating three clips (each clip consists of   four frames) from a skeleton sequence, 2)  using CNNs to learn   global long-term  temporal information of the skeleton sequence from each frame of  the generated clips, and 3)   using MTLN to 
 jointly train the CNN features of the four frames of the clips    to  incorporate the spatial structural information for action recognition.

 We also conducted the following baselines  to demonstrate the advantages of the proposed method:
    
 \textbf{Coordinates + FTP}~~In this baseline, the Fourier Temporal Pyramid (FTP)  \cite{wang2012mining} is applied to the 
   3D coordinates of the  skeleton sequences to extract   temporal features for action recognition.  This baseline
   is used to show the   benefits of using CNNs for long-term temporal modelling of the skeleton sequences.
 
  \textbf{Frames + CNN}~~In this baseline,  the CNN features of single frames instead of the entire generated clips are used for action recognition. In other words, only one   feature vector shown in Figure \ref{cnn1}(e) is used to train a neural network for classification.
  Thus the loss value of the network is given by Equation \ref{eql1}. 
 The average accuracy of the four features is provided. This baseline is
   used to show the benefits of using   the entire generated clips to incorporate the spatial structural information for action recognition.

 \textbf{Clips + CNN + Concatenation}~~In this baseline, the CNN features of all frames of the generated clips are concatenated before performing action recognition. In other words, the four  feature vectors shown in Figure \ref{cnn1}(e) are concatenated  and then fed to  a neural network for classification.  
 This baseline is used to show the benefits of using MTLN to process the features of the entire clips in parallel.
  
 \textbf{Clips + CNN + Pooling}~~In this baseline, max pooling is applied to the CNN features of all frames of the generate clips  before performing action recognition.
  Same as  Clips + CNN + Concatenation, this baseline is also used to show the benefits of using MTLN.

 \subsection{Datasets}

\textbf{NTU RGB+D Dataset}  \cite{Shahroudy_2016_CVPR}~~
 To   the best of our knowledge, this dataset is so far   the largest skeleton-based human action   dataset, with more than    56000 sequences  and 4 million frames. There are 60 classes of actions performed by 40 distinct subjects, including both one-person daily actions (e.g., clapping,
reading, 
writing) and two-person  interactions (\eg, handshaking, hug, pointing).    
These actions are captured by three cameras, which are placed at  different locations and view points. In total, there are 80 views for this dataset.  In this dataset, each skeleton has 25 joints.  The
3D coordinates of the  joints are provided.
Due to the large view point, intra-class and sequence length variations, the dataset is very challenging.

   \textbf{SBU Kinect Interaction Dataset}  \cite{yun2012two}~~
    This dataset was     collected using the Microsoft Kinect sensor. It contains  282 skeleton sequences and
6822 frames. In this dataset, each frame contains two persons performing an interaction. The interactions  include
approaching, departing, kicking, punching, pushing, hugging, shaking hands and exchanging. There are 15 joints for each skeleton.  This dataset is challenging due to the fact that the joint coordinates exhibit low accuracy \cite{yun2012two}. 
 
   \textbf{CMU Dataset}  \cite{cmu2013}~~
 This dataset contains   2235 sequences and about 1 million frames. For each skeleton, the 3D coordinates of 31 joints are provided. 
 The dataset has been categorized into 45 classes \cite{zhu2016co}. 
 All of the actions are performed by only one person. The dataset is   very challenging due to the large sequence length variations and intra-class diversity.

    \subsection{Implementation Details}

 For all datasets, the clips are generated with all frames of the original skeleton sequence without 
 any    pre-processing such as normalization, temporal down-sampling or noise filtering. 
 The proposed method was implemented using the MatConvNet
toolbox \cite{vedaldi2015matconvnet}. 
 The number of the  hidden unit of the first FC layer is set to  512.
 For the second FC layer (\ie, the output layer),  the number of the    unit is the same as the number of the action classes in each dataset.
  The network is trained using the  stochastic gradient
descent algorithm. The learning rate is set
to 0.001 and batch size is set to 100. The training is stopped
after 35 epochs. The performance of the proposed method on
each dataset is compared with existing methods using the
same testing protocol.

\subsection{Results}

\label{exp}

\textbf{NTU RGB+D Dataset}~~As in \cite{Shahroudy_2016_CVPR},  the evaluation on this dataset is performed with two standard protocols, \ie,   cross-subject evaluation and cross-view evaluation. 
In cross-subject evaluation, the sequences of   20 subjects are used for training and the data from  20   other subjects are used for testing. 
In cross-view evaluation, the sequences captured by two cameras are used for training and the rest  are used  for testing. 

The results are shown in Table \ref{ntu}. It can be seen that the proposed method  performs significantly better than others in both cross-subject and cross-view protocols.
The accuracy of the proposed method is   79.57\% when tested with the cross-subject protocol. Compared to the previous state-of-the-art method (ST-LSTM  + Trust Gate \cite{liu2016spatio}), the performance is improved by   10.37\%.
When tested with the cross-view protocol, the accuracy is improved from 77.7\% to 84.83\%.
 
The improved performance of the proposed method is due to the novel clip representation and feature learning method.
 As shown in   Table \ref{ntu}, 
  Frames + CNN 
   achieves an accuracy of about  75.73\% and 79.62\% for the two testing protocols, respectively.
   The performances are much better than Coordinates + FTP.   
 Compared to extracting temporal features of skeleton sequences with  FTP and native 3D coordinates, 
using CNN to learn the temporal information  of skeleton sequences from the generated frames  is    more robust  to noise and temporal variations due to the convolution and pooling operators, resulting in better performances.
    From  Table \ref{ntu}, it can also be seen that 
   Frames + CNN    also performs better than the previous state-of-the-art method. It clearly shows the effectiveness of the CNN features of the proposed clip representation. The performances are improved by learning entire clips with CNN and MTLN (\ie, Clips + CNN + MTLN). The improvements are about 4\% and 5\% for the two testing protocols, respectively. 
 It can also be seen that the proposed MTLN (\ie, Clips + CNN + MTLN) performs better than feature concatenation  (\ie, Clips + CNN + concatenation)  and pooling (\ie, Clips + CNN + pooling).
 Frames + CNN, Clips + CNN + concatenation and Clips + CNN + pooling can be viewed as a single-task method, while using MTLN to   process   multiple frames of the generated clips in parallel utilizes their intrinsic relationships and incorporates the spatial structural information, which improves the performance of the single-task method for   action recognition.  
 

\begin{table}[]
\centering
\tabcolsep=0.05cm
\caption {Performance on the NTU RGB+D dataset.}
\begin{tabular}{c|cc}
 
\hline
\multirow{2}{*}{Methods} & \multicolumn{2}{c}{Accuracy}\\
&Cross Subject&Cross View\\

\hline
Lie Group \cite{vemulapalli2014human}& 50.1\%&52.8\%\\
Skeletal Quads \cite{evangelidis2014skeletal}&38.6\%&41.4\%\\
Dynamic Skeletons \cite{hu2015jointly}&60.2\%&65.2\%\\
Hierarchical RNN \cite{du2015hierarchical}&59.1\%&64.0\%\\
Deep RNN \cite{Shahroudy_2016_CVPR}&59.3\%&64.1\%\\
Deep LSTM \cite{Shahroudy_2016_CVPR}&60.7\%&67.3\%\\
Part-aware LSTM \cite{Shahroudy_2016_CVPR}&62.9\%&70.3\%\\
ST-LSTM   \cite{liu2016spatio} & 65.2\%&76.1\%\\
ST-LSTM   + Trust Gate \cite{liu2016spatio} & 69.2\%&77.7\%\\
\hline
Coordinates + FTP&61.06\% &74.64\%\\
Frames + CNN & 75.73\%&79.62\%\\
Clips + CNN +  Concatenation  & 77.05\%&81.11\%\\
Clips + CNN +  Pooling  & 76.37\%&80.46\%\\
Clips + CNN +  MTLN  & \textbf{79.57\%}& \textbf{84.83\%}\\
\hline

\end{tabular}
 \vspace{-3mm}

\label{ntu}
\end{table}

 \textbf{SBU Kinect Interaction Dataset}~~
  As in \cite{yun2012two}, the evaluation of this dataset is a 5-fold cross validation, with the provided training/testing splits. Each frame of the skeleton sequences contains two   separate human skeletons. In this case,
  the two skeletons are  considered as two data samples and   the clip generation and   feature extraction are conducted separately for the two skeletons. For testing, the prediction of actions is obtained by averaging the    classification scores of the two samples. 
  
  Considering that the number of  samples in this dataset is too small, data augmentation is performed   to increase   the number of samples. More specifically, each frame image of the generated clips are  resized to  $250\times250$, and then random patches with size  of $224\times224$ are   cropped from the original image for feature learning using CNN.  For this dataset, 20 sub-images are cropped and the   total data samples are extended to 11320. 
 
 The comparisons  of the proposed method with other methods are shown  in Table \ref{sbu}. Similar to the NTU RGB+D dataset, CNN features perform  better than FTP to learn the temporal information. 
 It can be seen that when using  CNN features of individual frames, the accuracy is 90.88\%, which is similar to the    Deep LSTM + Co-occurrence method \cite{zhu2016co}. When incorporating the CNN features of the entire clips using concatenation and pooling methods,  the performance is improved by about 2\%. 
 The performance is improved to 93.57\%  when learning  the entire clips with MTLN.   It clearly shows the benefit
 of using MTLN to learn the CNN features entire clips.

 Since   the joint positions of this dataset are not very accurate  \cite{yun2012two},  existing methods including HBRNN \cite{du2015hierarchical} and Co-occurrence LSTM \cite{zhu2016co}  remove the joint noise by  smoothing the  position of    each joint using the Svaitzky-Golay   filter \cite{savitzky1964smoothing}. In \cite{liu2016spatio}, a  Trust Gate is introduced to remove the noisy joints and this improves the accuracy from 88.6\% to 93.3\%. 
Our method does not perform any pre-processing to handle   the noisy joints, but still performs better than all the others. It clearly shows that the   features learned from the generated clips are robust  to noise due to the convolution and pooling operators of the deep network.

\begin{table}[]
\centering
\tabcolsep=0.1cm
 
 \caption {Performance on the SBU kinect interaction dataset.}

\begin{tabular}{c|c}
\hline
Methods&Accuracy\\
\hline
\pbox{3cm} 
Raw Skeleton \cite{yun2012two} & 49.7\%\\
Joint Feature \cite{ji2014interactive}&86.9\%\\
CHARM \cite{li2015category} &83.9\%\\
Hierarchical RNN \cite{du2015hierarchical}&80.35\%\\
Deep LSTM \cite{zhu2016co}&86.03\%\\
Deep LSTM + Co-occurrence  \cite{zhu2016co}&90.41\%\\
ST-LSTM   \cite{liu2016spatio} &88.6\%\\
ST-LSTM + Trust Gate \cite{liu2016spatio} & 93.3\%\\
\hline 
Coordinates + FTP& 79.75\%\\
Frames + CNN & 90.88\%\\
Clips + CNN + Concatenation &92.86\% \\
Clips + CNN + Pooling &92.26\%\\
Clips + CNN + MTLN  & \textbf{93.57\%} \\
\hline

\end{tabular}
\label{sbu}
\end{table}

\begin{table}[t]
\centering
\tabcolsep=0.05cm
\caption {Performance on the CMU dataset.}
\begin{tabular}{c|cc}
\hline
\multirow{2}{*}{Methods} & \multicolumn{2}{c}{Accuracy}\\
&CMU subset&CMU\\

\hline
Hierarchical RNN  \cite{du2015hierarchical}&83.13\%&75.02\%\\
Deep LSTM \cite{zhu2016co}&86.00\%&79.53\%\\
Deep LSTM + Co-occurrence   \cite{zhu2016co}&88.40\%&81.04\%\\
\hline 
Coordinates + FTP &83.44\% &73.61\%\\
Frames + CNN  & 91.53\% &85.36\%\\
Clips + CNN + Concatenation &90.97\% &85.76\%\\ 
Clips + CNN + Pooling &90.66\%& 85.56\%\\
Clips + CNN+ MTLN  & \textbf{93.22\%}& \textbf{88.30\%}\\
\hline

\end{tabular}

\label{cmu1}
\end{table}

 \textbf{CMU Dataset}~~As in \cite{zhu2016co}, for this dataset, the evaluation is conducted on both the entire dataset with 2235 sequences, and a selected subset of 664 sequences. The subset includes 8 classes of actions, , \ie, basketball, cartwheel, getup, jump, pickup, run,
sit and walk back. 
 For the entire dataset, the testing protocol is  4-fold cross validation, and for the subset, it is evaluated 
 with 3-fold cross validation. The   training/tesing splits of the different folds are provided by \cite{zhu2016co}. 
 
 Similar to the SBU kinect interaction dataset,   data augmentation is also conducted on CMU dataset. For the entire dataset, each frame image  is used to generate 5 more images and the total data samples are extended   to 11175, and for the subset, the total samples are extended to 13280, which is 20 times of the original number. 
 
  The results are shown in  Table \ref{cmu1}. It can be seen that the performance of the proposed method is much better than   previous state-of-the-art methods on both the subset and the entire set. When    tested  on the subset, the accuracy of the proposed method was about 93.22\%, which is about 5\% better than the previous method \cite{zhu2016co}. 
   The performance on the entire dataset is improved from 81.04\% to  88.3\%. 
   
   \subsection{Discussions}

\textbf{Three gray clips or one color clip?}
As shown in Figure \ref{cnn1}, the frames of the three generated clips are gray images, each corresponding to  only one channel   of the cylindrical coordinates. Each frame is    duplicated three times to formulate a color image for CNN feature learning. The output CNN features of the three channels are concatenated in a feature vector for action recognition. A simple alternative is to generate a color clip  with three channels of the cylindrical coordinates, and then extract a single CNN feature from the color frame for action recognition. When this was tested on CMU dataset, the performance is 84.67\%, which is about 4\% worse than the proposed method. This is perhaps due to the fact that the relationship of the three generated channels is different from that of the RGB channels of natural color images.  The
RGB channels are arranged in sequence and   there is no
matching order between 3D coordinates and RGB channels.  
 
 \textbf{The more frames, the  better performance?}~~
 This paper uses only four reference joints   to generate clips, each having   four frames.  When 6 more joints are selected to generate more frames, \ie, the head, the left hand, the right hand, the left foot, the right foot  and the hip, the performance does not improve. When tested on CMU data, the performance is 86.01\%, which is about 2\% worse than the proposed method. This is due to the fact that the   other joints are not as stable as the selected four joints, which can   introduce noise.

\textbf{Cartesian coordinates or cylindrical coordinates?}~~
As mentioned in Section \ref{imagegene}, the 3D Cartesian coordinates of the vectors between the reference joints and the other joints are transformed to  cylindrical coordinates to generate clips.  We found that when using the original Cartesian coordinates for clip generation and action recognition, the performance drops. When tested on CMU dataset, the accuracy is 86.21\%, which is about 2\% worse than the proposed method. The cylindrical coordinates are more useful than the Cartesian coordinates to analyse the motions as each human skeleton utilizes pivotal joint movements to perform an action.

 \textbf{Features in different layers}~~
 As mentioned in Section \ref{cnnrep}, the feature maps in conv5\_1 layer  of the pre-trained CNN model  is adopted     as the representation of each input image.  We   found that using the features in the earlier layers  
 decreased the performance.  When using the features of   the conv4\_1 layer, the accuracy on CMU dataset is 84.59\%, which is about  4\% worse than the proposed method. This is perhaps due to the fact that the features in the earlier layers are not deep enough to capture the salient information of the input image.  
 We also found that   using the features in the later layers  
 made the    performance worse.
    When using the features of   the fc6 layer, the accuracy on CMU dataset is 83.52\%, which is about   5\% worse than the proposed method.
     This is because 
      the features   in the later layers  are more task-specific,  which largely rely on the original classes and dataset. The features of the later layers are thus less suitable than those of the earlier layers to transfer to other domains \cite{yosinski2014transferable, long2015learning}.

 \section{Conclusion}
 
 In this paper, we have proposed  to transform a skeleton sequence to three   video clips  for robust feature learning and action recognition. 
  We proposed to use   a pre-trained CNN model followed by a temporal pooling layer to extract a compact representation of each frame. The   CNN features of the three clips at the same time-step are concatenated in a single feature vector, 
 which describes the  temporal information of  the entire skeleton sequence and one particular spatial relationship between the joints. 
 We then propose an MTLN to jointly learn     the feature vectors at all the time-steps in parallel,     which   utilizes their intrinsic relationships and improves the performance for action recognition.
  We have tested the proposed method on three datasets, including NTU RGB+D dataset, SBU kinect interaction dataset and CMU dataset. Experimental results   have shown the effectiveness of the proposed new representation and feature learning method.
   
 \section{Acknowledgment}

This work was partially supported by Australian Research 
Council grants DP150100294, DP150104251, and DE120102960.
This paper used the NTU RGB+D Action Recognition Dataset made available by the ROSE Lab at the Nanyang Technological University, Singapore.







{\small
\bibliographystyle{ieee}
\bibliography{P1}

\begin{thebibliography}{10}\itemsep=-1pt

\bibitem{caruana1998multitask}
R.~Caruana.
\newblock Multitask learning.
\newblock In {\em Learning to learn}, pages 95--133. Springer, 1998.

\bibitem{chatfield2014return}
K.~Chatfield, K.~Simonyan, A.~Vedaldi, and A.~Zisserman.
\newblock Return of the devil in the details: Delving deep into convolutional
  nets.
\newblock {\em arXiv preprint arXiv:1405.3531}, 2014.

\bibitem{ciregan2012multi}
D.~Ciregan, U.~Meier, and J.~Schmidhuber.
\newblock Multi-column deep neural networks for image classification.
\newblock In {\em Computer Vision and Pattern Recognition (CVPR), 2012 IEEE
  Conference on}, pages 3642--3649. IEEE, 2012.

\bibitem{cmu2013}
CMU.
\newblock {CMU} graphics lab motion capture database.
\newblock In {\em http://mocap.cs.cmu.edu/}. 2013.

\bibitem{donahue2014decaf}
J.~Donahue, Y.~Jia, O.~Vinyals, J.~Hoffman, N.~Zhang, E.~Tzeng, and T.~Darrell.
\newblock Decaf: A deep convolutional activation feature for generic visual
  recognition.
\newblock In {\em International Conference on Machine Learning (ICML)}, pages
  647--655, 2014.

\bibitem{du2015hierarchical}
Y.~Du, W.~Wang, and L.~Wang.
\newblock Hierarchical recurrent neural network for skeleton based action
  recognition.
\newblock In {\em IEEE Conference on Computer Vision and Pattern Recognition
  (CVPR)}, pages 1110--1118, 2015.

\bibitem{evangelidis2014skeletal}
G.~Evangelidis, G.~Singh, and R.~Horaud.
\newblock Skeletal quads: Human action recognition using joint quadruples.
\newblock In {\em International Conference on Pattern Recognition (ICPR)},
  pages 4513--4518, 2014.

\bibitem{fernando2015modeling}
B.~Fernando, E.~Gavves, J.~M. Oramas, A.~Ghodrati, and T.~Tuytelaars.
\newblock Modeling video evolution for action recognition.
\newblock In {\em Proceedings of the IEEE Conference on Computer Vision and
  Pattern Recognition}, pages 5378--5387, 2015.

\bibitem{gaidon2013temporal}
A.~Gaidon, Z.~Harchaoui, and C.~Schmid.
\newblock Temporal localization of actions with actoms.
\newblock {\em IEEE transactions on pattern analysis and machine intelligence},
  35(11):2782--2795, 2013.

\bibitem{girshick2014rich}
R.~Girshick, J.~Donahue, T.~Darrell, and J.~Malik.
\newblock Rich feature hierarchies for accurate object detection and semantic
  segmentation.
\newblock In {\em IEEE Conference on Computer Vision and Pattern Recognition
  (CVPR)}, pages 580--587, 2014.

\bibitem{graves2012neural}
A.~Graves.
\newblock Neural networks.
\newblock In {\em Supervised Sequence Labelling with Recurrent Neural
  Networks}, pages 15--35. Springer, 2012.

\bibitem{graves2013speech}
A.~Graves, A.-r. Mohamed, and G.~Hinton.
\newblock Speech recognition with deep recurrent neural networks.
\newblock In {\em IEEE International Conference on Acoustics, Speech and Signal
  Processing}, pages 6645--6649. IEEE, 2013.

\bibitem{gu2016recurrent}
J.~Gu, G.~Wang, and T.~Chen.
\newblock Recurrent highway networks with language cnn for image captioning.
\newblock {\em arXiv preprint arXiv:1612.07086}, 2016.

\bibitem{han2016space}
F.~Han, B.~Reily, W.~Hoff, and H.~Zhang.
\newblock space-time representation of people based on 3d skeletal data: a
  review.
\newblock {\em arXiv preprint arXiv:1601.01006}, 2016.

\bibitem{han2015matchnet}
X.~Han, T.~Leung, Y.~Jia, R.~Sukthankar, and A.~C. Berg.
\newblock Matchnet: Unifying feature and metric learning for patch-based
  matching.
\newblock In {\em IEEE Conference on Computer Vision and Pattern Recognition
  (CVPR)}, pages 3279--3286, 2015.

\bibitem{hu2015jointly}
J.-F. Hu, W.-S. Zheng, J.~Lai, and J.~Zhang.
\newblock {Jointly learning heterogeneous features for RGB-D activity
  recognition}.
\newblock In {\em IEEE Conference on Computer Vision and Pattern Recognition
  (CVPR)}, pages 5344--5352, 2015.

\bibitem{hussein2013human}
M.~E. Hussein, M.~Torki, M.~A. Gowayyed, and M.~El-Saban.
\newblock Human action recognition using a temporal hierarchy of covariance
  descriptors on 3d joint locations.
\newblock In {\em IJCAI}, volume~13, pages 2466--2472, 2013.

\bibitem{ji2014interactive}
Y.~Ji, G.~Ye, and H.~Cheng.
\newblock Interactive body part contrast mining for human interaction
  recognition.
\newblock In {\em IEEE International Conference on Multimedia and Expo
  Workshops (ICMEW)}, pages 1--6. IEEE, 2014.

\bibitem{ke2017skeletonnet}
Q.~Ke, S.~An, M.~Bennamoun, F.~Sohel, and F.~Boussaid.
\newblock Skeletonnet: Mining deep part features for 3d action recognition.
\newblock {\em IEEE Signal Processing Letters}, 2017.

\bibitem{ke2016human}
Q.~Ke, M.~Bennamoun, S.~An, F.~Boussaid, and F.~Sohel.
\newblock Human interaction prediction using deep temporal features.
\newblock In {\em European Conference on Computer Vision Workshops}, pages
  403--414. Springer, 2016.

\bibitem{ke2014rotation}
Q.~Ke and Y.~Li.
\newblock Is rotation a nuisance in shape recognition?
\newblock In {\em Proceedings of the IEEE Conference on Computer Vision and
  Pattern Recognition}, pages 4146--4153, 2014.

\bibitem{koniusz2016tensor}
P.~Koniusz, A.~Cherian, and F.~Porikli.
\newblock Tensor representations via kernel linearization for action
  recognition from 3d skeletons.
\newblock {\em arXiv preprint arXiv:1604.00239}, 2016.

\bibitem{krizhevsky2012imagenet}
A.~Krizhevsky, I.~Sutskever, and G.~E. Hinton.
\newblock Imagenet classification with deep convolutional neural networks.
\newblock In {\em Advances in neural information processing systems}, pages
  1097--1105, 2012.

\bibitem{lecun1995convolutional}
Y.~LeCun, Y.~Bengio, et~al.
\newblock Convolutional networks for images, speech, and time series.
\newblock {\em The handbook of brain theory and neural networks},
  3361(10):1995, 1995.

\bibitem{li2015category}
W.~Li, L.~Wen, M.~Choo~Chuah, and S.~Lyu.
\newblock Category-blind human action recognition: A practical recognition
  system.
\newblock In {\em IEEE International Conference on Computer Vision (ICCV)},
  pages 4444--4452, 2015.

\bibitem{liu2016spatio}
J.~Liu, A.~Shahroudy, D.~Xu, and G.~Wang.
\newblock {Spatio-temporal LSTM with trust gates for 3D human action
  recognition}.
\newblock In {\em European Conference on Computer Vision (ECCV)}, pages
  816--833. Springer, 2016.

\bibitem{long2015learning}
M.~Long and J.~Wang.
\newblock Learning transferable features with deep adaptation networks.
\newblock {\em CoRR, abs/1502.02791}, 1:2, 2015.

\bibitem{nair2010rectified}
V.~Nair and G.~E. Hinton.
\newblock Rectified linear units improve restricted boltzmann machines.
\newblock In {\em International Conference on Machine Learning (ICML)}, pages
  807--814, 2010.

\bibitem{niebles2010modeling}
J.~C. Niebles, C.-W. Chen, and L.~Fei-Fei.
\newblock Modeling temporal structure of decomposable motion segments for
  activity classification.
\newblock In {\em European conference on computer vision}, pages 392--405.
  Springer, 2010.

\bibitem{pascanu2013construct}
R.~Pascanu, C.~Gulcehre, K.~Cho, and Y.~Bengio.
\newblock How to construct deep recurrent neural networks.
\newblock {\em arXiv preprint arXiv:1312.6026}, 2013.

\bibitem{peng2015encoding}
X.~Peng and C.~Schmid.
\newblock Encoding feature maps of cnns for action recognition.
\newblock 2015.

\bibitem{radenovic2016cnn}
F.~Radenovi{\'c}, G.~Tolias, and O.~Chum.
\newblock Cnn image retrieval learns from bow: Unsupervised fine-tuning with
  hard examples.
\newblock {\em arXiv preprint arXiv:1604.02426}, 2016.

\bibitem{sharif2014cnn}
A.~S. Razavian, H.~Azizpour, J.~Sullivan, and S.~Carlsson.
\newblock {CNN features off-the-shelf: an astounding baseline for recognition}.
\newblock In {\em IEEE Conference on Computer Vision and Pattern Recognition
  Workshops (CVPRW)}, pages 806--813, 2014.

\bibitem{russakovsky2015imagenet}
O.~Russakovsky, J.~Deng, H.~Su, J.~Krause, S.~Satheesh, S.~Ma, Z.~Huang,
  A.~Karpathy, A.~Khosla, M.~Bernstein, et~al.
\newblock Imagenet large scale visual recognition challenge.
\newblock {\em International Journal of Computer Vision}, 115(3):211--252,
  2015.

\bibitem{sainath2015convolutional}
T.~N. Sainath, O.~Vinyals, A.~Senior, and H.~Sak.
\newblock Convolutional, long short-term memory, fully connected deep neural
  networks.
\newblock In {\em Acoustics, Speech and Signal Processing (ICASSP), 2015 IEEE
  International Conference on}, pages 4580--4584. IEEE, 2015.

\bibitem{savitzky1964smoothing}
A.~Savitzky and M.~J. Golay.
\newblock Smoothing and differentiation of data by simplified least squares
  procedures.
\newblock {\em Analytical chemistry}, 36(8):1627--1639, 1964.

\bibitem{Shahroudy_2016_CVPR}
A.~Shahroudy, J.~Liu, T.-T. Ng, and G.~Wang.
\newblock {NTU RGB+D: A} large scale dataset for {3D} human activity analysis.
\newblock In {\em IEEE Conference on Computer Vision and Pattern Recognition
  (CVPR)}, June 2016.

\bibitem{simonyan2014very}
K.~Simonyan and A.~Zisserman.
\newblock Very deep convolutional networks for large-scale image recognition.
\newblock {\em arXiv preprint arXiv:1409.1556}, 2014.

\bibitem{szegedy2015going}
C.~Szegedy, W.~Liu, Y.~Jia, P.~Sermanet, S.~Reed, D.~Anguelov, D.~Erhan,
  V.~Vanhoucke, and A.~Rabinovich.
\newblock Going deeper with convolutions.
\newblock In {\em Proceedings of the IEEE Conference on Computer Vision and
  Pattern Recognition}, pages 1--9, 2015.

\bibitem{vedaldi2015matconvnet}
A.~Vedaldi and K.~Lenc.
\newblock Matconvnet: Convolutional neural networks for matlab.
\newblock In {\em ACM International Conference on Multimedia}, pages 689--692,
  2015.

\bibitem{veeriah2015differential}
V.~Veeriah, N.~Zhuang, and G.-J. Qi.
\newblock Differential recurrent neural networks for action recognition.
\newblock In {\em IEEE International Conference on Computer Vision (ICCV)},
  pages 4041--4049, 2015.

\bibitem{vemulapalli2014human}
R.~Vemulapalli, F.~Arrate, and R.~Chellappa.
\newblock Human action recognition by representing 3d skeletons as points in a
  lie group.
\newblock In {\em IEEE Conference on Computer Vision and Pattern Recognition
  (CVPR)}, pages 588--595, 2014.

\bibitem{wang2012mining}
J.~Wang, Z.~Liu, Y.~Wu, and J.~Yuan.
\newblock Mining actionlet ensemble for action recognition with depth cameras.
\newblock In {\em IEEE Conference on Computer Vision and Pattern Recognition
  (CVPR)}, pages 1290--1297, 2012.

\bibitem{wang2014latent}
L.~Wang, Y.~Qiao, and X.~Tang.
\newblock Latent hierarchical model of temporal structure for complex activity
  classification.
\newblock {\em IEEE Transactions on Image Processing}, 23(2):810--822, 2014.

\bibitem{wang2016temporal}
L.~Wang, Y.~Xiong, Z.~Wang, Y.~Qiao, D.~Lin, X.~Tang, and L.~Van~Gool.
\newblock Temporal segment networks: towards good practices for deep action
  recognition.
\newblock In {\em European Conference on Computer Vision}, pages 20--36.
  Springer, 2016.

\bibitem{wang2016action}
P.~Wang, Z.~Li, Y.~Hou, and W.~Li.
\newblock Action recognition based on joint trajectory maps using convolutional
  neural networks.
\newblock In {\em Proceedings of the 2016 ACM on Multimedia Conference}, pages
  102--106. ACM, 2016.

\bibitem{weinland2006free}
D.~Weinland, R.~Ronfard, and E.~Boyer.
\newblock Free viewpoint action recognition using motion history volumes.
\newblock {\em Computer vision and image understanding}, 104(2):249--257, 2006.

\bibitem{weston2014memory}
J.~Weston, S.~Chopra, and A.~Bordes.
\newblock Memory networks.
\newblock {\em arXiv preprint arXiv:1410.3916}, 2014.

\bibitem{xia2012view}
L.~Xia, C.-C. Chen, and J.~Aggarwal.
\newblock View invariant human action recognition using histograms of {3D}
  joints.
\newblock In {\em IEEE Conference on Computer Vision and Pattern Recognition
  Workshops (CVPRW)}, pages 20--27, 2012.

\bibitem{xiong2015recognize}
Y.~Xiong, K.~Zhu, D.~Lin, and X.~Tang.
\newblock Recognize complex events from static images by fusing deep channels.
\newblock In {\em Proceedings of the IEEE Conference on Computer Vision and
  Pattern Recognition}, pages 1600--1609, 2015.

\bibitem{yang2012eigenjoints}
X.~Yang and Y.~L. Tian.
\newblock Eigenjoints-based action recognition using
  naive-bayes-nearest-neighbor.
\newblock In {\em IEEE Computer Society Conference on Computer Vision and
  Pattern Recognition Workshops (CVPRW)}, pages 14--19, 2012.

\bibitem{yosinski2014transferable}
J.~Yosinski, J.~Clune, Y.~Bengio, and H.~Lipson.
\newblock How transferable are features in deep neural networks?
\newblock In {\em Advances in neural information processing systems}, pages
  3320--3328, 2014.

\bibitem{yun2012two}
K.~Yun, J.~Honorio, D.~Chattopadhyay, T.~L. Berg, and D.~Samaras.
\newblock Two-person interaction detection using body-pose features and
  multiple instance learning.
\newblock In {\em IEEE Conference on Computer Vision and Pattern Recognition
  Workshops (CVPRW)}, pages 28--35, 2012.

\bibitem{zhu2016co}
W.~Zhu, C.~Lan, J.~Xing, W.~Zeng, Y.~Li, L.~Shen, and X.~Xie.
\newblock Co-occurrence feature learning for skeleton based action recognition
  using regularized deep lstm networks.
\newblock In {\em AAAI Conference on Artificial Intelligence (AAAI)}, 2016.

\end{thebibliography}
}

\end{document}